\documentclass{article}

\usepackage{microtype}
\usepackage{graphicx}
\usepackage{subfigure}
\usepackage{booktabs} 
\usepackage{caption}

\usepackage{hyperref}

\usepackage[accepted]{icml2025}

\usepackage{amsmath}
\usepackage{amssymb}
\usepackage{mathtools}
\usepackage{amsthm}
\usepackage{subcaption}
\usepackage{colortbl} 
\usepackage[capitalize,noabbrev]{cleveref}

\theoremstyle{plain}

\theoremstyle{definition}

\theoremstyle{remark}

\usepackage[textsize=tiny]{todonotes}

\icmltitlerunning{PIAD-SRNN}

\begin{document}

\twocolumn[
\icmltitle{PIAD-SRNN: Physics-Informed Adaptive Decomposition in State-Space RNN 
}



\icmlsetsymbol{equal}{*}

\begin{icmlauthorlist}
\icmlauthor{Ahmad Mohammadshirazi}{yyy}
\icmlauthor{Pinaki Prasad Guha Neogi}{yyy}
\icmlauthor{Rajiv Ramnath}{yyy}
\end{icmlauthorlist}

\icmlaffiliation{yyy}{Department of Computer Science and Engineering, Ohio State University, Ohio, US}

\icmlcorrespondingauthor{Ahmad Mohammadshirazi}{mohammadshirazi.2@osu.edu}

\icmlkeywords{Long-Term and Short-Term Time Series Forecasting, State-Space, Physics-based Equations, Recurrent Neural Networks, Decomposition, Missing Value Imputation}

    

\vskip 0.3in
]



\printAffiliationsAndNotice{\icmlEqualContribution} 

\begin{abstract}
Time series forecasting often demands a trade-off between accuracy and efficiency. While recent Transformer models have improved forecasting capabilities, they come with high computational costs. 
Linear-based models have shown better accuracy than Transformers but still fall short of ideal performance.
We propose PIAD-SRNN—a physics-informed adaptive decomposition state-space RNN—that separates seasonal and trend components and embeds domain equations in a recurrent framework. 
We evaluate PIAD-SRNN's performance on indoor air quality datasets, focusing on CO$_2$ concentration prediction across various forecasting horizons, and results demonstrate that it consistently outperforms SoTA models in both long-term and short-term time series forecasting, including transformer-based architectures, in terms of both MSE and MAE.
Besides proposing PIAD-SRNN which balances accuracy with efficiency, 
this paper also provides four curated datasets. Code and data: https://github.com/ahmad-shirazi/DSSRNN


\end{abstract}

\vspace{-6mm}
\section{Introduction}
\label{sec:intro}

Time series forecasting demands both expressiveness and efficiency while also robustly handling missing data. On the one hand, self‐attention forecasting models (e.g., Transformers) capture long‐range dependencies but incur quadratic compute and tuning overhead; conversely, linear‐decomposition methods run fast but miss nonlinear dynamics \cite{zeng2023transformers}. In addition, existing imputation approaches (for missing values in real-time data) 
either ignore system insights or add significant computational cost.



To address all these gaps in one step, we introduce the Physics‐Informed Adaptive Decomposition State‐Space Recurrent Neural Network (PIAD‐SRNN). This idea was inspired by the success of nature-inspired~\cite{8821993, 10.1007/978-981-15-2188-1_29} and physics-oriented~\cite{10.1007/978-981-15-5616-6_21, kar2020triangular} modeling paradigms in ML and optimization problems. PIAD‐SRNN embeds physics‐based state‐space equations within a recurrent architecture that decomposes each time series into trend and seasonal components. By merging mechanistic modeling with data‐driven decomposition (inspired by DLinear \cite{zeng2023transformers}, FEDformer~\cite{zhou2022fedformer}), PIAD‐SRNN captures both global patterns and fine‐grained temporal variations while retaining linear‐time complexity per step.
We validate PIAD‐SRNN on four Indoor Air Quality (IAQ) datasets we created, focusing on CO$_2$ concentration forecasting across multiple horizons (up to 720 steps). Compared to leading Transformer and linear baselines, PIAD-SRNN achieves substantial gains in MSE and MAE, while requiring minimal inference time and memory. Our contributions are:
{\bf (1) Model.} We propose PIAD‐SRNN, a novel architecture that integrates physics‐informed state‐space modeling with adaptive decomposition for efficient and accurate long-term and short-term time series forecasting and imputation, and is also robust to outliers.  
{\bf (2) Datasets.} We release four curated IAQ datasets and all code to foster reproducibility and advance research in physics‐informed time series learning.
 {\bf (3) Evaluation.} Through extensive experiments 
 we demonstrate that PIAD‐SRNN consistently outperforms 
 SoTA baselines in both predictive accuracy as well as computational efficiency.

  


\section{Related Work}
\label{sec:related}

\subsection{Time Series Imputation}
\vspace{-2mm}
Time series imputation has evolved from traditional RNN-based methods like GRU-D \cite{che2018recurrent}, BRITS \cite{cao2018brits}, and M-RNN \cite{yoon2018estimating}, which incorporate time decay factors and bidirectional processing but struggle with long-term dependencies and computational intensity. Generative approaches using GANs and VAEs \cite{luo2018multivariate, liu2019naomi, fortuin2020gp} offer sophisticated distribution learning but face training complexity and scalability issues. Recent self-attention-based models like SAITS \cite{du2023saits} optimize imputation through diagonally-masked attention blocks, though computational efficiency remains challenging.

\subsection{Time Series Forecasting}

Classical statistical methods like ARIMA \cite{siami2018comparison} and \cite{hyndman2018forecasting} excel at capturing linear relationships and explicit seasonal patterns in univariate time series, but they face limitations with nonlinear dynamics and complex multivariate relationships.
Transformer architectures \cite{vaswani2017attention} revolutionized forecasting through self-attention mechanisms, spawning specialized variants: Informer \cite{zhou2021informer} addresses quadratic complexity with ProbSparse attention, Autoformer \cite{wu2021autoformer} employs Auto-Correlation mechanisms, and iTransformer \cite{liu2023itransformer} processes variate tokens for multivariate modeling. Surprisingly, DLinear \cite{zeng2023transformers} demonstrated that simple linear decomposition can outperform complex Transformers, challenging the necessity of architectural complexity. Recent approaches like SegRNN \cite{lin2023segrnn} uses segment-wise iterations, \& TiDE \cite{das2023long} combines linear models with MLPs.

PIAD-SRNN model integrates physics-informed state-space modeling with decomposition techniques from DLinear and Autoformer, achieving superior computational efficiency while maintaining the interpretability and robustness required for real-time environmental applications. For a detailed literature review, see Appendix~\ref{sec:app-related}.

\section{Data Preparation}
\label{sec:data_imputation}

\subsection{Dataset Description}
The scarcity of IAQ datasets and their limited availability present a significant challenge to research in this domain. For our study, we created datasets from Lawrence Berkeley National Laboratory's Building 59 in California, covering the period from Aug. 19, 2019, to Dec. 31, 2021, creating 20,760 data points from different office settings. 
It encompasses hourly indoor CO$_2$ measurements along with other crucial environmental variables—a collection of data infrequently encountered in existing studies. The location of each office numbered 44, 45, 62, and 68 within the building is detailed in a map provided in \cite{luo2022three}, offering contextual insight into the data collection environment. A key aspect of our research is the prediction of indoor CO$_2$ concentration ($CO_{2\_in}$), using five current environmental variables such as current indoor CO$_2$ levels, time of day (Hour), day of the week ($\text{Num\_Week}$), and both indoor and outdoor temperatures ($T_{\text{in}}$ and $T_{\text{out}}$) \cite{mohammadshirazi2022predicting, mohammadshirazi2023novel}. Appendix~\ref{sec:app-data-analysis} presents data analysis \& validations showing complexities of the real-world datasets.

\subsection{Handling Missing Observations}
Table \ref{tab:inputvar} provides a detailed examination of the data absence across various input variables collected from four distinct office environments. 
To effectively manage this dataset, we segregated it into training, validation, and testing subsets, with all instances of missing data allocated to the testing subset.
The proposed PIAD-SRNN model embeds physics-informed state-space equations within an adaptive seasonal–trend decomposition, enabling it to perform imputation and forecasting in a single recurrent framework.  We fit the model by minimizing MSE and MAE on the validation set—first recovering missing CO$_2$ values, then seamlessly sliding into multi-horizon prediction (and outlier clarification-Appendix~\ref{app-4}).  The injected physical priors yield robustness to substantial data gaps, ensuring reliable IAQ reconstruction and long-term accuracy.  
Appendix~\ref{sec:outlier-class} shows the detection and distribution of outliers in our datasets.

\begin{table}

\begin{center}
 \scalebox{.83}{

\begin{tabular}{l |ccc} 
 \hline
 Input Variables & $CO_{2-in}$  & $T_{in}$ & $T_{out}$ \\
 \hline
 Office1 & 2,823 (13.6\%)  & 3,027 (14.6\%) & 2,823 (13.6\%) \\ 
 Office2,3,4 & 2,823 (13.6\%) & 2,823 (13.6\%) & 2,823 (13.6\%) \\
 \hline
\end{tabular}
}
\caption{Input Variables with Missing Entries for Predicting Indoor CO$_2$ Concentrations across Four Office Environments}
\label{tab:inputvar}
\end{center}
\end{table}





\section{Physics‐Informed Recurrent Modeling}
\label{sec:model_arch}

Here, we present two nested architectures. First, \textbf{PI‐SRNN} (Physics‐Informed State‐Space RNN) embeds a classical CO$_2$ mass‐balance into an RNN; then \textbf{PIAD‐SRNN}  builds on PI‐SRNN by splitting each series into trend and seasonal components for improved multi‐horizon forecasting.

\subsection{Physics‐Informed State‐Space RNN (PI‐SRNN)}  
The PI‐SRNN architecture represents a novel blend of traditional state-space models with the dynamic capabilities of recurrent neural networks (RNNs). At its core, PI‐SRNN aims to model time series data by capturing both the underlying physical processes and the temporal dependencies. This is achieved through a combination of state variables that embody the physical state of the system, and RNN layers that process sequential data, allowing for the integration of prior knowledge and temporal context. The PI‐SRNN architecture is designed to efficiently handle the flow of information through time, making it particularly adept at predicting complex dynamics in time series data, such as those encountered in IAQ monitoring. 
The transformation from physics to ML equations begins with the foundational state-space representation of indoor CO$_2$ dynamics \cite{persily2022development}, as shown in Equation \ref{eq:CO$_2$_equation}. 

\begin{equation}
\frac{\partial }{\partial t}CO_{2\_in} = \frac {\dot m }{\rho V} (CO_{2\_out} - CO_{2\_in}) + \frac {M_{t} }{\rho}
\label{eq:CO$_2$_equation}
\end{equation}

This equation models the temporal change in indoor CO$_2$ concentration by relating it to variables that capture both the physical characteristics of the environment and the dynamics of CO$_2$ exchange. Specifically, the mass flow rate $\dot{m}$ reflects the air exchange due to ventilation, represented by \(\frac{\dot{m}}{\rho V} \times CO_{2\_out}\), $\rho$ signifies air density, $V$ stands for the volume of the indoor space, and $M_t$ represents the internal generation of CO$_2$, such as those generated by occupants or appliances within the space. The variable $CO_{2\_out}$, meanwhile, captures the concentration of CO$_2$ outside the building. Together, these variables form the basis of our state-space approach, setting the stage for its translation into an ML framework.

Equation \ref{eq:CO$_2$_equation2} is where the conversion to an ML framework takes place. The term $A$ represents the state matrix, encapsulating the system dynamics, and $B$ the input matrix, incorporating external influences. These matrices interact with the state vector $S_t$ (Equation \ref{eq:CO$_2$_equation3}) and input vector $U_t$ (Equation \ref{eq:CO$_2$_equation4}) to form a discretized version of the physics-based model, respectively. This interaction is key to defining the subsequent state vector $S_{t+1}$ and the differential state change $dS_{t+1}$. These state vectors, embedded within the ML architecture, allow neural networks to approximate the intricate relationships found within the dataset. The continuous state update from $S_t$ to $S_{t+1}$ (Equation \ref{eq:CO$_2$_equation3}), facilitated by the differential state $dS_{t+1}$, is a discrete analog of the physical process. This reformulation allows us to leverage the power of neural networks to approximate the complex relationships within the data, retaining the interpretability and structure provided by the physical model.





\begin{figure}[t!]
  \centering
  \includegraphics[width=0.8\columnwidth]{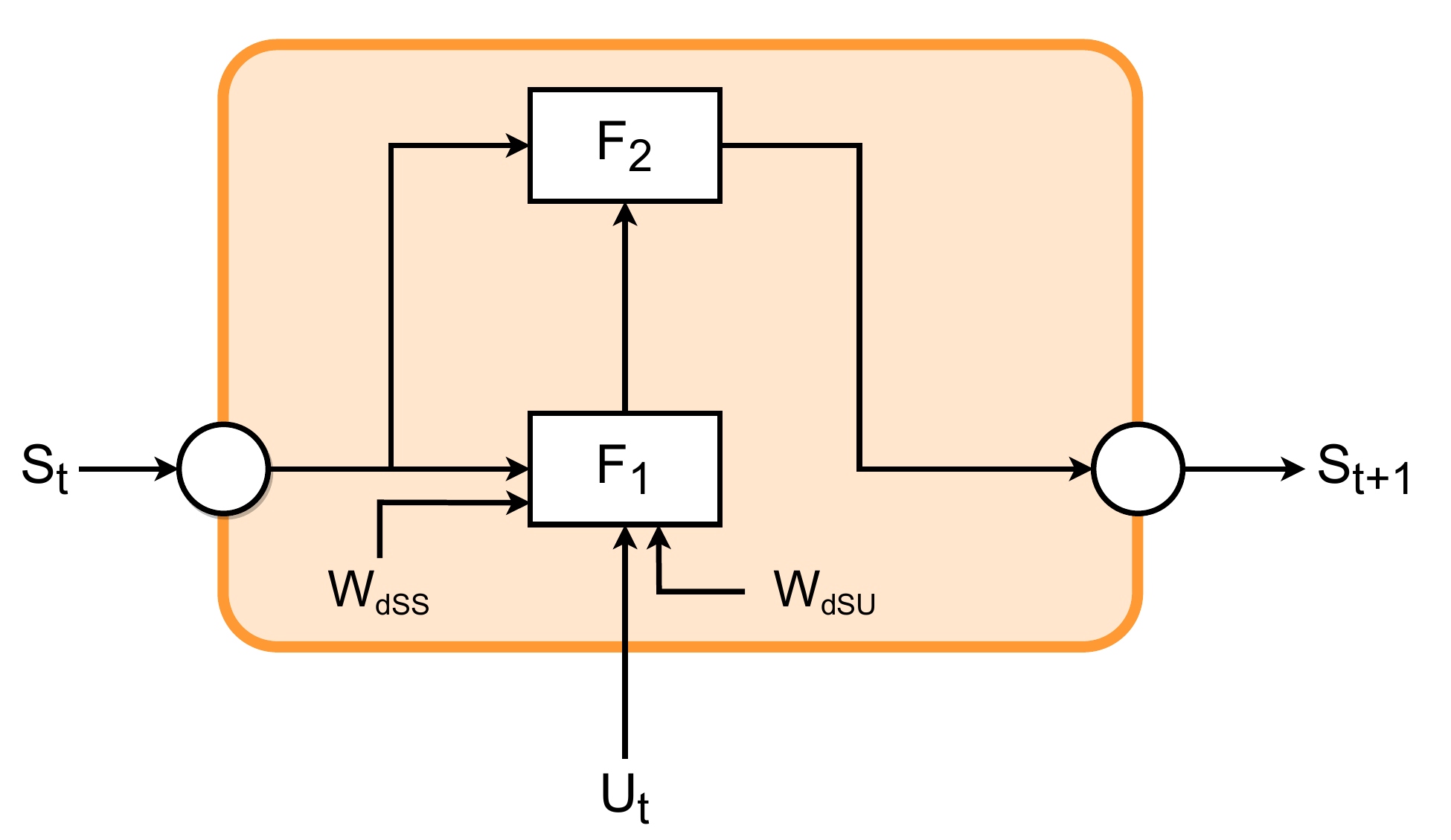}
  \caption{Physics-Informed State-Space RNN (PI-SRNN).}
  \label{fig:SSRNN}
\end{figure}


\begin{equation}
\scalebox{0.82}{$
\underbrace{\frac{\partial }{\partial t}(CO_{2\_in})}_{dS_{t+1}} = 
\underbrace{- \frac{\dot m }{\rho V}}_{A} \times 
\underbrace{CO_{2\_in}}_{S_{t}} 
+ \left\{ 
\underbrace{\frac{\dot m }{\rho V} \times CO_{2\_out} + \frac{M_{t}}{\rho}}_{B {\times} {U_{t}}} 
\right\}
$}
\label{eq:CO$_2$_equation2}
\end{equation}

\begin{equation}
S_{t+1} \leftarrow  {CO_{2\_in_{t+1}}} , S_{t} \leftarrow  {CO_{2\_in_{t}}}
\label{eq:CO$_2$_equation3}
\end{equation}


\begin{equation}
U_{t} \leftarrow  {CO_{2\_in_{t}}}, {Hour}_{t}, T_{in_{t}}, T_{out_{t}}, Num\_week_{t} 
\label{eq:CO$_2$_equation4}
\end{equation}


The PI-SRNN model, presented in Figure \ref{fig:SSRNN}, is meticulously crafted to process inputs within each discrete time step. The model assimilates the current state vector $S_t$ and the input vector $U_t$ to advance its internal state. The transformation of $S_t$ alongside $U_t$ is governed by a nonlinear function $F_1$, resulting in the interim state change $dS_{t+1}$, as shown in equation \ref{eq:CO$_2$_equation5}. This intermediate state is then sculpted by another nonlinear function $F_2$, culminating in the updated state $S_{t+1}$ (Equation \ref{eq:CO$_2$_equation6}). Weight matrices $W_{dSS}$ and $W_{dSU}$, alongside bias vectors $b_{dSS}$ and $b_{dSU}$, underpin this evolution, modulating the influence of past and present information and embedding non-linearity via ReLU activation functions, denoted as $F_1$ and $F_2$ (see equation \ref{eq:CO$_2$_equation7}).

Based on Figure \ref{fig:SSRNN} representation, the PI-SRNN model state space and its changes are  updated as follows:  


\begin{equation}
dS_{t+1} \leftarrow F_{1}(S_{t} \times W_{dSS} + b_{dSS} + U_{t} \times W_{dSU} + b_{dSU})
\label{eq:CO$_2$_equation5}
\end{equation}


\begin{equation}
S_{t+1}  \leftarrow F_{2}(S_{t} +  dS_{t+1})
\label{eq:CO$_2$_equation6}
\end{equation}


\begin{equation}
\scalebox{0.81}{$
S_{t+1}  \leftarrow F_{2}(S_{t} + F_{1}(S_{t} \times W_{dSS} + b_{dSS} + U_{t} \times W_{dSU} + b_{dSU})
$}
\label{eq:CO$_2$_equation7}
\end{equation}


It should be noted that the equation referenced in \ref{eq:CO$_2$_equation}, while primarily developed for modeling CO$_2$, can be adapted to encompass a variety of indoor pollutants such as NO$_2$, O$_3$, PM$_1$, PM$_{2.5}$, and PM$_{10}$. This adaptation necessitates modifying the generation term $M_t$ for each specific pollutant's source strength and incorporating particular deposition or removal processes. Such modifications allow the PI-SRNN model to extend its applicability across a range of indoor pollutants, demonstrating its versatility in different environmental conditions. In the following subsection, we will expound on the intricacies of the PIAD-SRNN and its implementation for enhanced predictive performance.

\begin{figure}[t!]
  \includegraphics[scale=0.34]{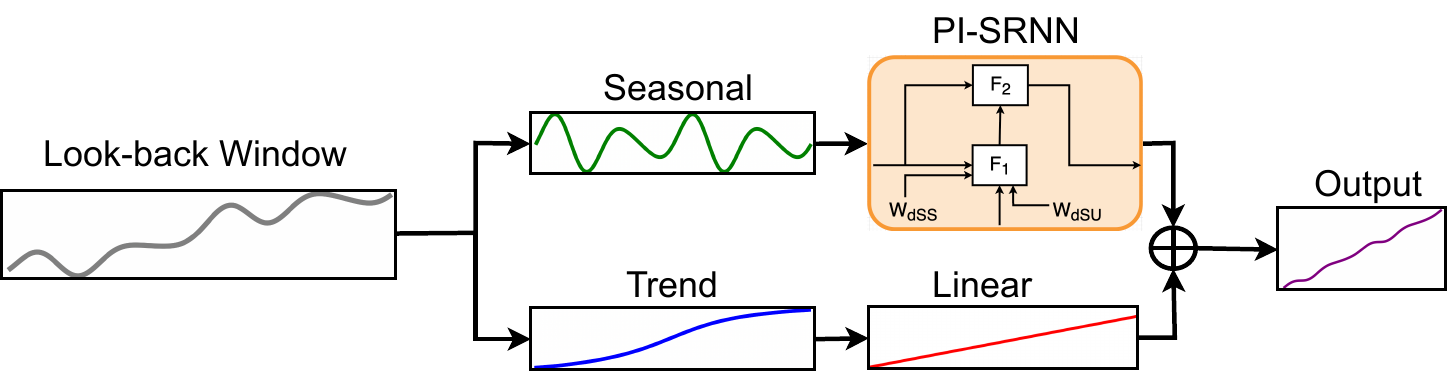}
  \caption{Physics-Informed Adaptive Decomposition State-Space Recurrent Neural Network (PIAD-SRNN).}
  \label{fig:PIAD-SRNN}
\end{figure}

\begin{table*}[]
\centering
\setlength{\tabcolsep}{2pt}
\scalebox{0.76}{
\begin{tabular}{c|c|cccc|cccc|cccc|cccc} 
\cline{1-18}
Methods & Metric & \multicolumn{4}{c|}{Office 1 (44)} & \multicolumn{4}{c|}{Office 2 (45)} & \multicolumn{4}{c|}{Office 3 (62)} & \multicolumn{4}{c}{Office 4 (68)} \\ 
\cline{3-18}
 &  & T = 96 & T = 192 & T = 336 & T = 720 & T = 96 & T = 192 & T = 336 & T = 720 & T = 96 & T = 192 & T = 336 & T = 720 & T = 96 & T = 192 & T = 336 & T = 720  \\ 
 \cline{1-18}

Informer & MSE & 0.531 &  0.567 & 0.514 & 0.625 & 0.523 & 0.555 & 0.549 & 0.593 & 0.522 & 0.535 & 0.510 & 0.529 & 0.576 & 0.577 & 0.522 & 0.520  \\
 & MAE & 0.508 & 0.527 & \underline{0.482} & 0.553 & 0.504 & 0.519 & 0.509 & 0.538 & 0.477 & 0.471 & 0.465 & 0.484 & 0.503 & 0.516 & 0.502 & \underline{0.492}   \\  

 \cline{1-18}
PatchTST & MSE & \underline{0.431} & \underline{0.484} & \underline{0.497} & \underline{0.526} & \underline{0.410} & \underline{0.461} & \underline{0.475} & \underline{0.496} & \underline{0.382} & \underline{0.442} & \underline{0.456} & \underline{0.491} & \underline{0.422} & \underline{0.484} & \underline{0.501} & \underline{0.516} \\
 & MAE & \underline{0.460} & \underline{0.485} & 0.489 & \underline{0.504} & \underline{0.452} & \underline{0.476} & \underline{0.480} & \underline{0.489} & \underline{0.420} & \underline{0.453} & \underline{0.460} & \underline{0.473} & \underline{0.453} & \underline{0.484} & \underline{0.491} & 0.495 \\
\hline

  \hline 
  \hline

TiDE & MSE & 0.491 & 0.521 & 0.557 & 0.612 & 
        0.481 & 0.526 & 0.573 & 0.648 & 
        \underline{0.391} & \underline{0.404} & \underline{0.411} & \underline{0.436} & 
        0.507 & 0.536 & 0.561 & 0.613 \\
      & MAE & 0.471 & 0.489 & 0.502 & 0.535 &
        0.473 & 0.490 & 0.510 & 0.549 & 
        \underline{0.415} & \underline{0.422} & \underline{0.432} & \underline{0.447} & 
        0.488 & 0.500 & 0.513 & 0.541 \\


\cline{1-18}
{Linear} & MSE & \underline{0.452} & 0.503 & 0.508 & 0.536 & 0.426 & 0.477 & 0.483 & 0.502 & 0.397 & 0.450 & 0.463 & 0.497 & 0.436 & 0.495 & 0.507 & 0.519  \\
 & MAE & \underline{0.475} & 0.499 & 0.501 & 0.509 & 0.462 & 0.487 & 0.489 & 0.498 & 0.438 & 0.466 & 0.475 & 0.488 & 0.469 & 0.498 & 0.504 & 0.502 \\ 

 
\cline{1-18}
{DLinear} & MSE & 0.454 & \underline{0.500} & \underline{0.504} & \underline{0.535} & \underline{0.422} & \underline{0.474} & \underline{0.479} & \underline{0.499} & 0.392 & 0.446 & 0.458 & 0.494 & \underline{0.429} & \underline{0.491} & \underline{0.503} & \underline{0.517}  \\
 & MAE & 0.475 & \underline{0.497} & \underline{0.498} & \underline{0.508} & \underline{0.459} & \underline{0.485} & \underline{0.486} & \underline{0.496} & 
 0.434 & 0.463 & 0.472 & 0.485 & \underline{0.464} & \underline{0.495} & \underline{0.501} & \underline{0.501}  \\

 \hline 
  \hline

\cline{1-18}
SegRNN & MSE & 0.491 & 0.564 & 0.554 & 0.576 & 0.483 & 0.571 & 0.560 & 0.544 & 0.453 & 0.549 & 0.548 & 0.580 & 0.509 & 0.600 & 0.630 & 0.613 \\
       & MAE & 0.488 & 0.523 & 0.520 & 0.530 & 0.486 & 0.524 & 0.518 & 0.513 & 0.462 & 0.506 & 0.509 & 0.521 & 0.504 & 0.548 & 0.566 & 0.558 \\

\cline{1-18}
{PI-SRNN} & MSE & 0.393 & 0.444 & 0.450 & 0.477 & 0.370 & 0.418 & 0.424 & 0.444 & 0.342 & 0.391 & 0.403 & 0.438 & 0.378 & 0.437 & 0.448 & 0.460 \\
& MAE & 0.416 & 0.440 & 0.442 & 0.450 & 0.406 & 0.428 & 0.430 & 0.439 & 0.382 & 0.406 & 0.416 & 0.429 & 0.410 & 0.439 & 0.444 & 0.443  \\ 
\hline

\rowcolor{blue!20}{PIAD-SRNN} & MSE & \textbf{0.378} & \textbf{0.428} & \textbf{0.434} & \textbf{0.463} & \textbf{0.353} & \textbf{0.403} & \textbf{0.409} & \textbf{0.429} & \textbf{0.320} & \textbf{0.375} & \textbf{0.388} & \textbf{0.425} & \textbf{0.359} & \textbf{0.421} & \textbf{0.433} & \textbf{0.445} \\
\rowcolor{blue!20}& MAE & \textbf{0.401} & \textbf{0.426} & \textbf{0.428} & \textbf{0.437} & \textbf{0.390} & \textbf{0.415} & \textbf{0.416} & \textbf{0.426} & \textbf{0.362} & \textbf{0.392} & \textbf{0.402} & \textbf{0.416} & \textbf{0.394} & \textbf{0.424} & \textbf{0.431} & \textbf{0.430}
  \\ 

  \hline
\end{tabular}}

\caption{Comparative Forecasting Performance: This table presents the MSE and MAE across forecasting horizons (T) of {96, 192, 336, 720} for four office environments, with input sequence length L=96. Bold figures indicate the best overall performance, while underlined figures show the top results for transformer-based, MLP-based, and linear-based models. (More comparisons in Appendix~\ref{sec:app-a}-Table \ref{tab:appendix-results})}
\label{tab:res-pred}
\end{table*}

\subsection{Physics‐Informed Adaptive Decomposition (PIAD‐SRNN)}  
Although PI‐SRNN effectively blends domain knowledge with sequence learning, complex IAQ patterns benefit from separating slow trends from fast cycles.  PIAD‐SRNN applies a 24 h moving average to each series, yielding:
\[
X_t = T_t + S_t,
\]
where $T_t$ (trend) and $S_t$ (seasonal) feed into two parallel paths (Fig.~\ref{fig:PIAD-SRNN}):  
\textbf{ (1) Trend branch:} a lightweight linear layer captures smooth, long‐term shifts with minimal computation. 
\textbf{(2) Seasonal branch:} a PI‐SRNN block models cyclical fluctuations via its physics‐informed recurrent updates.
The final prediction is simply
\[
\widehat{X}_{t+h} \;=\; \widehat{T}_{t+h} \;+\; \widehat{S}_{t+h}\,,
\]
This regularizes learning across time scales and reduces interference between components.  Empirically, PIAD‐SRNN yields superior MSE/MAE on multi‐horizon IAQ benchmarks while maintaining low inference cost.


\vspace{-3mm}

\section{Experiments}
\label{sec:result}

In the results section of our paper, we evaluate the proposed models using the robust NVIDIA A100 GPU at the Ohio Supercomputer Center (OSC) \footnote{https://www.osc.edu/}. The forthcoming subsections will present a detailed account of the model's imputation and prediction capabilities, each scrutinized through rigorous experimentation. These findings will illustrate PIAD-SRNN's strengths in accurately forecasting and handling missing data within the context of IAQ assessment.


\subsection{Imputation Analysis}
\label{sec:imputation_analysis}


Table~\ref{result-imput} compares single-step (T=1) imputation performance of PIAD-SRNN with SoTA techniques like DLinear~\cite{zeng2023transformers}, and SAITS~\cite{du2023saits}. PIAD-SRNN consistently achieves the lowest MSE and MAE across all 4 offices, underscoring its exceptional ability to reconstruct missing IAQ measurements with unparalleled precision. By embedding physics-driven state-space dynamics within an adaptive seasonal–trend decomposition, PIAD-SRNN captures complex temporal patterns and robustly handles data sparsity. This evaluation demonstrates that our model’s advanced architecture is instrumental in learning and imputing data patterns, outshining SoTA imputation methods.

\begin{table}[ht]
\centering
\setlength{\tabcolsep}{5pt}
\scalebox{.86}{
\begin{tabular}{c|c|cccc} 
\hline
Method       & Metric & Office 1 & Office 2 & Office 3 & Office 4 \\
\hline
\rowcolor{blue!20}PIAD-SRNN    & MSE    & \textbf{0.357} & \textbf{0.180} & \textbf{0.217} & \textbf{0.373} \\
             \rowcolor{blue!20}& MAE    & \textbf{0.375} & \textbf{0.284} & \textbf{0.293} & \textbf{0.367} \\ 
\hline
DLinear      & MSE    & 0.383 & 0.199 & 0.233 & 0.397 \\
             & MAE    & 0.393 & 0.299 & 0.304 & 0.382 \\  
\hline
SAITS        & MSE    & 1.079 & 0.956 & 0.922 & 0.954   \\
             & MAE    & 0.686 & 0.701 & 0.669 & 0.710   \\ 
\hline
\end{tabular}
}
\caption{One-step imputation (T=1) performance: MSE and MAE for PIAD-SRNN, DLinear, and SAITS across four offices. Bold entries denote best results.}
\label{result-imput}
\end{table}

\subsection{Prediction Analysis}
\label{sec:prediction}

We benchmark PIAD-SRNN against SoTA baselines following Zeng et al.~\cite{zeng2023transformers} for multi-horizon forecasting at $T\!=\!\{96,192,336,720\}$. Table \ref{tab:res-pred} (and Appendix~\ref{sec:app-a}) shows that PIAD-SRNN achieves the lowest MSE/MAE across all horizons and office environments, outperforming DLinear—the top linear model—as well as Transformer (listed in Table~\ref{tab:appendix-results}) at shorter horizons and Informer \cite{zhou2021informer} at longer horizons.

We also evaluate the impact of the look-back window size for both long-term (720 steps) and short-term (24 steps) forecasting in Appendix~\ref{app-b}-Fig.~\ref{fig:STSF}.
PIAD-SRNN consistently balances accuracy and efficiency, outperforming high-cost models like TiDE \cite{das2023long} in most settings. These results underscore PIAD-SRNN’s adaptability across temporal scales and its superior accuracy–efficiency trade-off. 




\subsection{Predicting outlier events} 
\label{sec:outlier-class}

\begin{table}[ht]
\centering
\scalebox{.87}{
\begin{tabular}{l|ccc}
\hline
 & {Threshold}  & {Number of events} & {Percentage} \\
 & & {Training/(Testing)} & {Training/(Testing)} \\
\hline
Office 1 & 451.12 & 1112/(268) & 7.70\%/(6.46\%) \\
Office 2  & 450.84 & 1019/(260) & 7.06\%/(6.26\%) \\
Office 3 & 450.75 & 1175/(292) & 8.14\%/(7.03\%) \\
Office 4 & 453.05 & 928/(271) & 6.43\%/(6.53\%) \\
\hline
\end{tabular}
}
\caption{Comparative analysis across training and testing datasets between the number of high-concentration CO\textsubscript{2} events and the corresponding percentage of data points for each office alongside their third quartile threshold}

\label{tab:event-method}
\end{table}

In order to demonstrate the robustness of our model in predicting outlier events within the CO$_2$ data, we employ a binary classification approach based on the box-and-whisker plot method. Specifically, CO$_2$ measurements above the upper whisker—calculated as $Q_3 + 1.5 \times IQR$ where $Q_3$ is the third quartile and $IQR$ is the interquartile range—are classified as outliers and encoded as '1', while non-outliers are assigned '0' (See the threshold value for each office in Table \ref{tab:event-method}. This binary transformation allows us to specifically focus on the prediction of outlier occurrences. We then compare the performance of our model in predicting these binary outcomes against SoTA models, showcasing our model's capability to identify significant deviations in air quality. The number of outlier events detected for each office is systematically cataloged in Table \ref{tab:event-method}. The comparative results and analysis are presented in detail in the Result section of the paper.

\subsection{Additional Model Performance Analysis}
\label{sec:app-a}

\begin{table*}[!htbp]
\centering
\setlength{\tabcolsep}{2.5pt}
\scalebox{0.74}{
\begin{tabular}{c|c|cccc|cccc|cccc|cccc} 
\cline{1-18}
Methods & Metric & \multicolumn{4}{c|}{Office 1 (44)} & \multicolumn{4}{c|}{Office 2 (45)} & \multicolumn{4}{c|}{Office 3 (62)} & \multicolumn{4}{c}{Office 4 (68)} \\ 
\cline{3-18}
 &  & T = 96 & T = 192 & T = 336 & T = 720 & T = 96 & T = 192 & T = 336 & T = 720 & T = 96 & T = 192 & T = 336 & T = 720 & T = 96 & T = 192 & T = 336 & T = 720  \\ 
 \cline{1-18}
ARIMA & MSE & 0.881 & 0.887 & 1.398 & 1.359 & 1.208 & 1.022 & 1.320 & 1.452 & 1.064 & 0.939 & 1.194 & 1.191 & 1.076 & 0.949 & 1.442 & 1.561 \\
 & MAE & 0.640 & 0.729 & 0.862 & 0.838 & 0.845 & 0.808 & 0.858 & 0.872 & 0.735 & 0.764 & 0.801 & 0.773 & 0.751 & 0.766 & 0.930 & 0.911 \\
 
 \cline{1-18}
ETS & MSE & 0.910 & 0.904 & 1.435 & 1.387 & 1.181 & 1.007 & 1.301 & 1.436 & 1.008 & 0.910 & 1.146 & 1.149 & 0.960 & 0.893 & 1.313 & 1.433 \\
 & MAE & 0.655 & 0.736 & 0.877 & 0.847 & 0.831 & 0.801 & 0.850 & 0.866 & 0.706 & 0.753 & 0.780 & 0.762 & 0.700 & 0.748 & 0.879 & 0.874 \\

  \hline 
  \hline
\cline{1-18}
FEDformer & MSE & 1.062 & 1.208 & 1.166 & 1.186 & 1.008 & 1.054 & 0.974 & 1.143 & 1.070 & 1.076 & 0.944 & 0.891 & 1.030 & 1.137 & 1.123 & 0.878 \\
 & MAE & 0.740 & 0.792 & 0.770 & 0.786 & 0.723 & 0.740 & 0.709 & 0.768 & 0.745 & 0.749 & 0.709 & 0.691 & 0.750 & 0.793 & 0.788 & 0.687 \\
\cline{1-18}
Transformer & MSE & 0.528 & 0.599 & 0.651 & 0.643 & 0.421 & 0.555 & 0.572 & 0.553 & 0.471 & 0.636 & 0.500 & 0.534 & 0.528 & 0.529 & 0.713 & 0.581  \\
 & MAE & 0.497 & 0.536 & 0.566 & 0.565 & 0.461 & 0.539 & 0.533 & 0.525 & 0.448 & 0.522 & 0.465 & 0.477 & 0.497 & 0.490 & 0.566 & 0.525  \\
\cline{1-18}
Autoformer & MSE & 0.619 & 0.667 & 0.767 & 0.704 & 0.586 & 0.613 & 0.603 & 0.599 & 0.537 & 0.607 & 0.934 & 0.620 & 0.648 & 0.729 & 1.193 & 0.645\\
 & MAE & 0.598 & 0.583 & 0.648 & 0.598 & 0.556 & 0.557 & 0.561 & 0.561 & 0.527 & 0.555 & 0.698 & 0.555 & 0.591 & 0.629 & 0.809 & 0.586 \\ 
 
\cline{1-18}
iTransformer & MSE & 0.562 & 0.520 & 0.585 & 0.606 & 0.433 & 0.483 & 0.509 & 0.619 & 0.562 & 0.520 & 0.585 & 0.606 & 0.555 & 0.574 & 0.659 & 0.713 \\
 & MAE & 0.463 & 0.460 & 0.486 & 0.488 & 0.459 & 0.484 & 0.439 & 0.484 & 0.463 & 0.463 & 0.486 & 0.488 & 0.472 & 0.482 & 0.496 & 0.511 \\



\cline{1-18}
ModernTCN & MSE & 0.491 & 0.586 & 0.586 & 0.605 & 0.460 & 0.549 & 0.549 & 0.551 & 0.428 & 0.522 & 0.531 & 0.567 & 0.473 & 0.586 & 0.615 & 0.599 \\
          & MAE & 0.482 & 0.525 & 0.526 & 0.529 & 0.474 & 0.516 & 0.513 & 0.512 & 0.445 & 0.490 & 0.495 & 0.506 & 0.487 & 0.541 & 0.556 & 0.543 \\


\cline{1-18}
{Nlinear} & MSE & 0.513 & 0.615 & 0.611 & 0.621 & 0.476 & 0.569 & 0.569 & 0.569 & 0.448 & 0.547 & 0.557 & 0.585 & 0.494 & 0.614 & 0.642 & 0.616   \\
 & MAE & 0.500 & 0.545 & 0.542 & 0.542 & 0.485 & 0.527 & 0.523 & 0.522 & 0.462 & 0.506 & 0.510 & 0.518 & 0.501 & 0.556 & 0.571 & 0.554   \\ 
 

 \hline 
  \hline

VanillaRNN & MSE & 0.452 & 0.508 & 0.498 & 0.544 & 0.411 & 0.463 & 0.499 & 0.526 & 0.399 & 0.439 & 0.477 & 0.477 & 0.432 & 0.513 & 0.485 & 0.524 \\
& MAE & 0.470 & 0.512 & 0.493 & 0.505 & 0.450 & 0.465 & 0.495 & 0.508 & 0.428 & 0.457 & 0.452 & 0.471 & 0.462 & 0.499 & 0.481 & 0.501 \\

\cline{1-18}
GRU & MSE & 0.451 & 0.490 & 0.528 & 0.558 & 0.512 & 0.542 & 0.518 & 0.586 & 0.459 & 0.453 & 0.481 & 0.504 & 0.467 & 0.523 & 0.549 & 0.526 \\
& MAE & 0.459 & 0.474 & 0.483 & 0.503 & 0.498 & 0.514 & 0.502 & 0.546 & 0.446 & 0.456 & 0.472 & 0.476 & 0.484 & 0.508 & 0.520 & 0.503 \\

\cline{1-18}
LSTM & MSE & 0.456 & 0.507 & 0.538 & 0.548 & 0.422 & 0.494 & 0.538 & 0.538 & 0.420 & 0.427 & 0.447 & 0.479 & 0.454 & 0.556 & 0.584 & 0.516 \\
& MAE & 0.456 & 0.480 & 0.499 & 0.497 & 0.453 & 0.500 & 0.526 & 0.529 & 0.438 & 0.443 & 0.457 & 0.472 & 0.477 & 0.520 & 0.536 & 0.494 \\

  \hline
\end{tabular}}
\caption{Additional Model Comparisons: Performance metrics for supplementary models across different prediction horizons. These results complement the primary findings presented in Table \ref{tab:res-pred}.}
\label{tab:appendix-results}
\end{table*}

To provide a comprehensive evaluation of our PIAD-SRNN model, we conducted additional comparative analyses with both traditional and modern time series forecasting approaches. Table \ref{tab:appendix-results} presents an extended comparison of performance metrics across different models and prediction horizons for all four office environments. Among the traditional statistical models, ARIMA and ETS demonstrate consistent but relatively higher error rates across all prediction horizons. For instance, in Office 1, ARIMA shows MSE values ranging from 0.881 to 1.398, while ETS exhibits similar performance patterns with MSE values between 0.910 and 1.435. These results highlight the limitations of traditional statistical approaches in capturing complex temporal dependencies in indoor air quality data. The neural network-based models (VanillaRNN, GRU, and LSTM) show improved performance compared to statistical methods. Notably, GRU and LSTM achieve competitive results, particularly for shorter prediction horizons (T=96), with MSE values consistently below 0.5 across all offices. This improvement can be attributed to their ability to learn and retain temporal patterns in the data. FEDformer and Transformer architectures demonstrate varying degrees of success, with Transformer showing particularly strong performance in shorter prediction horizons but experiencing some degradation in accuracy for longer forecasting periods. The Autoformer model maintains relatively stable performance across different prediction lengths, though not achieving the optimal results of our proposed PIAD-SRNN model. The Nlinear model, while computationally efficient, shows moderate performance levels that remain relatively consistent across different prediction horizons, suggesting its utility in scenarios where computational resources are constrained.

These supplementary results further validate the effectiveness of our PIAD-SRNN approach, as presented in the manuscript's Table \ref{tab:res-pred}, by providing additional context and comparison points across a broader range of methodologies and architectural approaches.

\subsection{Extended Visual Analysis of Model Performance}
\label{app-b}
\begin{figure*}[!htbp]
    \centering
    
    \subfigure[\textbf{24} steps - \textbf{Office 1}]{
        \includegraphics[width=0.31\textwidth]{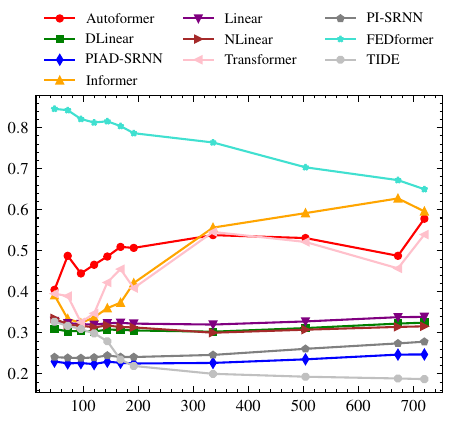}
        \label{fig:sub1}
    }
    \hfill
    \subfigure[\textbf{24} steps - \textbf{Office 2}]{
        \includegraphics[width=0.31\textwidth]{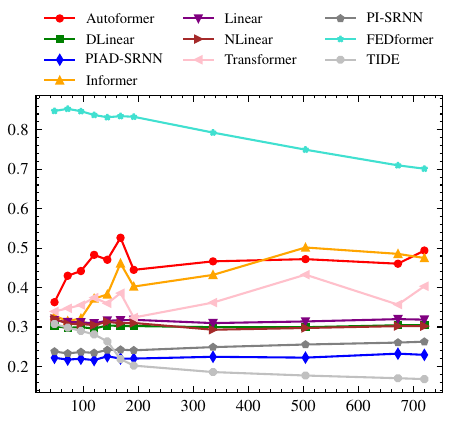}
        \label{fig:sub2}
    }
    \hfill
    \subfigure[\textbf{24} steps - \textbf{Office 3}]{
        \includegraphics[width=0.31\textwidth]{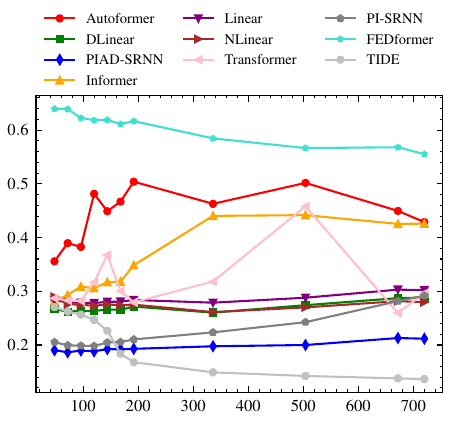}
        \label{fig:sub3}
    }
    
    \subfigure[\textbf{24} steps - \textbf{Office 4}]{
        \includegraphics[width=0.31\textwidth]{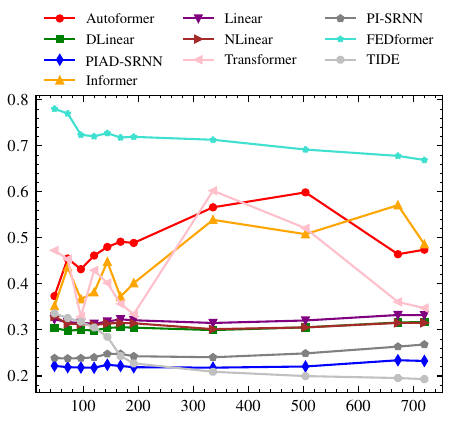}
        \label{fig:sub4}
    } 
\hfill
    \subfigure[\textbf{720} steps - \textbf{Office 1}]{
        \includegraphics[width=0.31\textwidth]{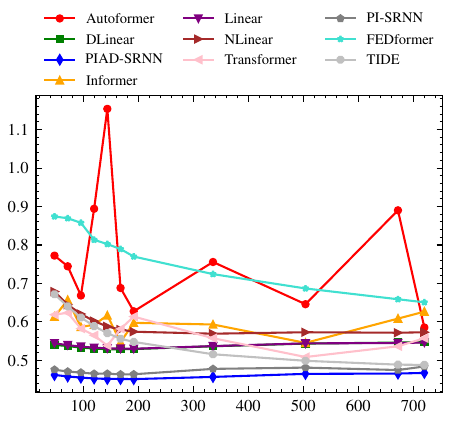}
        \label{fig:sub5}
    }
    \hfill
    \subfigure[\textbf{720} steps - \textbf{Office 2}]{
        \includegraphics[width=0.31\textwidth]{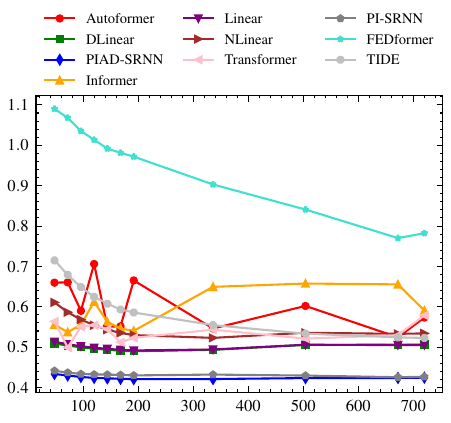}
        \label{fig:sub6}
    }
    \hfill
    \subfigure[\textbf{720} steps - \textbf{Office 3}]{
        \includegraphics[width=0.31\textwidth]{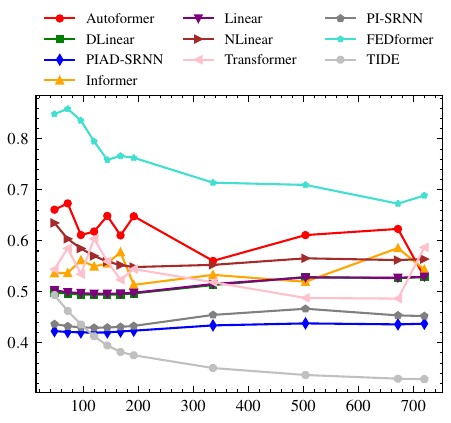}
        \label{fig:sub7}
    }
    \hfill
    \subfigure[\textbf{720} steps - \textbf{Office 4}]{
        \includegraphics[width=0.31\textwidth]{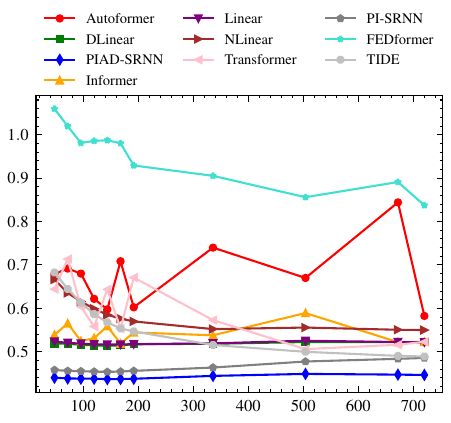}
        \label{fig:sub8}
    }

    \caption{Comparative MSE Analysis for Forecasting Models: Illustrated here are the MSE metrics for models with varying look-back window sizes for both long-term (e.g., 720 steps) and short-term (e.g., 24 steps) forecasting. The Y-axis measures MSE, while the X-axis represents the look-back window size, highlighting performance across four offices.}
    \label{fig:STSF}
\end{figure*}

Figure \ref{fig:STSF} provides a comprehensive visualization of model performance across different temporal scales and office environments. The analysis is structured to compare short-term (24 steps) and long-term (720 steps) forecasting capabilities, offering insights into how various models handle different prediction horizons. In the short-term forecasting scenario (24 steps), shown in the top row of Figure \ref{fig:STSF}, we observe that PIAD-SRNN maintains consistently lower MSE values across all four offices compared to other models. This is particularly evident in Office 1 and Office 3, where the performance gap between PIAD-SRNN and other models becomes more pronounced as the look-back window size increases. The Transformer and Autoformer models show competitive performance but exhibit more volatility in their error rates, especially with larger look-back windows. The long-term forecasting results (720 steps), displayed in the bottom row, reveal the robustness of PIAD-SRNN in maintaining prediction accuracy over extended time horizons. While most models show degraded performance with increased prediction length, PIAD-SRNN demonstrates remarkable stability. This is particularly notable in Office 2 and Office 4, where other models show significant performance deterioration with larger look-back windows.
The graphs also highlight an interesting pattern regarding the impact of look-back window size on model performance. For most models, including TIDE and FEDformer, increasing the look-back window beyond certain thresholds (typically around 300-400 steps) does not yield substantial improvements in prediction accuracy. In contrast, PIAD-SRNN shows more consistent improvement with increased historical data, suggesting better utilization of temporal information.

These visualizations support our quantitative findings presented in Table \ref{tab:res-pred} of the manuscript, providing additional evidence for PIAD-SRNN's superior performance in both short-term and long-term forecasting scenarios. The consistent performance across different offices and temporal scales demonstrates the robustness and reliability of our proposed approach.

\subsection{Computational Efficiency}
\label{app-3}

\begin{table}
\scalebox{.93}{
\begin{tabular}{l|llll}
\hline
Method      & MACs   & Parameter & Time    & Memory  \\ \hline
PIAD-SRNN      & 0.11G  & 408.21K   & 0.58ms  & 437MiB  \\ 
DLinear     & 0.04G  & 139.7K    & 0.15ms  & 196MiB  \\ 
TiDE        & 1.09G  & 2.69M     & 32.87ms & 1011MiB \\ 
PatchTST    & 4.79G  & 18.21M    & 14.03ms & 2033MiB \\ 
Transformer & 4.03G  & 13.61M    & 10.31ms & 1740MiB \\ 
Informer    & 3.93G  & 14.39M    & 18.96ms & 1105MiB \\ 
FEDformer   & 4.41G  & 20.68M    & 15.58ms & 1183MiB \\ \hline
\end{tabular}}
\caption{Comparative analysis of computational efficiency between PIAD-SRNN, DLinear, and various transformer-based models, evaluated on the Office \#1 air quality dataset with input sequence length \(L=96\) and prediction sequence length \(T=720\). Metrics include multiply-accumulate operations (MACs), model parameters, average inference time over five runs, and memory usage.}
\label{tab:performance}
\end{table}

In our comparative analysis, delineated in Table \ref{tab:performance}, we scrutinize the computational efficiency of our proposed PIAD-SRNN model compared to DLinear and a suite of Transformer-based architectures. This evaluation is predicated on the Office \#1 air quality dataset, adopting an input sequence length \(L=96\) and a prediction sequence length \(T=720\). The criteria for this comparison encompass multiply-accumulate operations (MACs), the aggregate number of model parameters, the average inference time over a quintet of runs, and the requisite memory allocation during computational processes.

DLinear distinguished itself as the most computationally efficient model, demanding the least computational resources with a mere 0.04G MACs and minimal memory usage of 196MiB, alongside a remarkable average inference speed of 0.15ms. This exceptional efficiency renders it particularly advantageous for resource-constrained settings. Close on its heels, PIAD-SRNN stands as the second most efficient model, consuming slightly more resources at 0.11G MACs and 437MiB of memory but still delivering a swift inference time of 0.58ms, validating its efficacy.

\subsection{Classification}
\label{app-4}

\begin{table}[!h]
\centering
\setlength{\tabcolsep}{4pt}
\scalebox{0.81}{
\begin{tabular}{l|l|cccc}
\hline
{Performance } & & & & & \\
Metrics & Model &  Office 1 & Office 2 & Office 3 & Office 4 \\
 
\hline
{True Positive}& PIAD-SRNN & 192 & 209 & 234 & 218 \\
& DLinear & 181 & 189 & 223 & 203 \\
& PatchTST & 179 & 178 & 219 & 207 \\
\hline
{False Positive}& PIAD-SRNN  & 84 & 80 & 46 & 67 \\
& DLinear & 137 & 111 & 67 & 95 \\
& PatchTST &  98 & 99 & 57 & 98 \\
\hline
{True Negative}& PIAD-SRNN  & 3,799 & 3,811 & 3,813 & 3,807 \\
& DLinear & 3,757 & 3,780 & 3,792 & 3,785 \\
& PatchTST & 3,785 & 3,792 & 3,802 & 3,782 \\
\hline
{False Negative}& PIAD-SRNN  & 76 & 51 & 46 & 73 \\
& DLinear & 87 & 71 & 69 & 68 \\
& PatchTST & 89 & 82 & 73 & 64 \\
\hline
{Accuracy}& PIAD-SRNN  & 96.15\% & 96.84\% & 97.52\% & 96.96\% \\
& DLinear & 94.87\% & 95.62\% & 96.72\% & 96.07\% \\
& PatchTST &  95.50\% & 95.64\% & 96.87\% & 96.10\% \\
\hline
{Precision}& PIAD-SRNN  & 69.57\% & 72.32\% & 84.63\% & 74.91\% \\
& DLinear & 58.96\% & 63.00\% & 76.90\% & 68.12\% \\
& PatchTST &  64.62\% & 64.26\% & 79.35\% & 67.87\% \\
\hline
{Recall}& PIAD-SRNN  & 71.64\% & 80.38\% & 80.48\% & 80.44\% \\
& DLinear & 67.54\% & 72.69\% & 76.37\% & 74.91\%  \\
& PatchTST &  66.79\% & 68.46\% & 75.00\% & 76.38\% \\
\hline
{F1 Score}& PIAD-SRNN  & 70.59\% & 76.14\% & 82.02\% & 77.58\% \\
& DLinear & 62.96\% & 67.50\% & 76.63\% & 71.35\%  \\
& PatchTST &  65.69\% & 66.29\% & 77.11\% & 71.88\% \\
\hline
\end{tabular}}

\caption{Comparative analysis of model performance metrics for $CO_{2}$ concentration event classification over T = 96.}
\label{tab: res-clas}
\end{table}

In an effort to further comprehend the regression capabilities of our PIAD-SRNN model, we transformed the regression task into a classification problem, as meticulously explained in Section \ref{sec:outlier-class}. The results of this transformation are depicted in Table \ref{tab: res-clas}, where we delineate the thresholds set for event detection, the count of events, and their respective percentages for each office environment.

The classification approach allows us to dissect the model's performance in distinguishing between high-concentration and normal event occurrences. While the DLinear model exhibits commendable recall for high-concentration events—likely a result of its decomposition capability, breaking down time series into seasonal and trend components—the Autoformer model shows a preference for normal events. This tendency is attributed to its Transformer-based architecture, which is proficient in modeling long-range dependencies within the data.

Notwithstanding the strengths displayed by DLinear and Autoformer in their respective domains, the PIAD-SRNN model demonstrates a clear overarching superiority. It transcends the typical decomposition approach by integrating state space physics concepts, offering a profound understanding of the dynamics of the time series data. PIAD-SRNN consistently maintains higher True Positive rates alongside lower False Positive rates, indicating an adeptness at correctly identifying events while minimizing false alarms—a critical aspect for real-world applications. The True Negative rate is also the highest for PIAD-SRNN, reflecting its strong discriminative power in differentiating non-events.
Precision and Recall are metrics where PIAD-SRNN distinctly outshines its counterparts. The model's precision is indicative of the reliability and exactitude of its predictions, while the higher recall rates—especially for high-concentration events—demonstrate an increased sensitivity and a lower propensity to overlook significant events.

\section{Conclusion and Future Work}
\label{sec:conlcusion}

This study introduced the Physics-Informed Adaptive Decomposition State-Space Recurrent Neural Network (PIAD-SRNN), which combines physics-based modeling with state-space methodologies for accurate time-series forecasting and data imputation, demonstrating superior computational efficiency and outperforming transformer-based models in both MSE and MAE to deliver a robust solution for environmental data analysis. Future work will extend PIAD-SRNN to broader applications, including energy consumption forecasting and climate-control optimization, thereby validating its generalizability across diverse domain-specific contexts.

\nocite{langley00}

\bibliography{example_paper}
\bibliographystyle{icml2025}

\newpage
\appendix

\clearpage


\section*{Appendix}

\section{Related Work (Detailed)}
\label{sec:app-related}

We survey background work in two areas: imputation of missing time‐series data and multi‐horizon forecasting.

\subsection{Imputation Methods}
Accurate imputation in time series is crucial for dataset preparation, as in our case detailed in Table \ref{tab:inputvar}. Traditional RNN-based methods, such as GRU-D \cite{che2018recurrent}, have introduced time decay factors to address missing data. Approaches like BRITS \cite{cao2018brits} and M-RNN \cite{yoon2018estimating} enhance this with bidirectional RNNs, yet they grapple with the inherent complexities of long-term dependencies and computational intensity. Generative models, including GANs and VAEs, provide alternative strategies \cite{luo2018multivariate, liu2019naomi, casale2018gaussian, fortuin2020gp}, but their training complexity and interpretability issues limit their practical scalability.

The emergence of self-attention-based models has introduced a new paradigm in time series imputation. Models like DeepMVI \cite{bansal2021missing} and NRTSI \cite{shan2023nrtsi} leverage Transformer architectures for handling multidimensional and irregularly sampled time series data. Despite their novel approach, these models face challenges in computational efficiency and generalizability. Du et al. (2023) recently advanced this field with SAITS, a model that optimizes imputation through diagonally-masked self-attention blocks, addressing both temporal dependencies and feature correlations with enhanced efficiency \cite{du2023saits}.

Our approach diverges significantly from these existing methods. As noted in the introduction, we integrate domain knowledge with ML. This enhances the accuracy of our model while also addressing the limitations of existing approaches.  Additionally, leveraging neural networks within this framework allows for efficient processing of larger datasets, balancing robustness with computational agility. 

\subsection{Forecasting Techniques}
Time series forecasting has deep roots in statistical methods. The Autoregressive Integrated Moving Average (ARIMA) model \cite{siami2018comparison} has been a cornerstone of time series analysis since the 1970s, decomposing time series into autoregressive and moving average components while handling non-stationarity through differencing. Error, Trend, and Seasonality (ETS) models \cite{hyndman2018forecasting} provide another classical approach, explicitly modeling these three components with various combinations of additive and multiplicative effects. While these methods excel at capturing linear relationships and explicit seasonal patterns in univariate time series, they face limitations with nonlinear dynamics and complex multivariate relationships.

Transformer models by Vaswani et al. (2017) \cite{vaswani2017attention}, have revolutionized time series forecasting through their innovative use of self-attention mechanisms, enabling the model to focus selectively on relevant parts of the input. This model's parallelizable architecture significantly enhances its efficiency in handling large datasets. There are also newer models based on transformers such as PatchTST \cite{nie2022time}, Informer \cite{zhou2021informer}, and Autoformer \cite{wu2021autoformer}.

Furthermore, Informer model by Zhou et al. (2021) \cite{zhou2021informer}, which addresses the quadratic time complexity and memory usage inherent in traditional Transformer models, introduces the ProbSparse self-attention mechanism, a distilling operation, and a generative decoder to manage long-sequence time-series forecasting effectively. This model demonstrates a marked improvement in prediction capabilities for long-sequence forecasting tasks by mitigating the limitations of the encoder-decoder architecture.

The FEDformer model, introduced by Zhou et al. (2022) \cite{zhou2022fedformer}, represents another advancement in time series forecasting by integrating Transformer architecture with seasonal-trend decomposition. This hybrid approach enables the FEDformer to capture both the global trends and the intricate details within the time series data. By leveraging the sparsity in the frequency domain, FEDformer achieves enhanced efficiency and effectiveness in long-term forecasting, outpacing conventional Transformer models in both performance and computational efficiency.

In contrast, the Autoformer model by Wu et al. (2021) \cite{wu2021autoformer} incorporates an Auto-Correlation mechanism and a novel series decomposition strategy within its architecture. This approach allows the Autoformer to dissect and analyze complex time series data, focusing on seasonal patterns while progressively incorporating trend components. This model's unique design facilitates a deeper understanding of temporal dependencies, particularly in sub-series levels, thereby offering a robust solution for long-term forecasting challenges.

iTransformer \cite{liu2023itransformer} introduces a perspective by inverting the traditional transformer architecture. Rather than processing temporal tokens, it operates on variate tokens, enabling more effective modeling of multivariate correlations. This approach maintains the advantages of transformer architectures while addressing their limitations in time series contexts, particularly for long sequences and complex multivariate relationships.

In addition, the introduction of a linear decomposition model, Dlinear, by Zeng et al. (2023) \cite{zeng2023transformers} challenges the prevailing dominance of Transformer-based models in long-term time series forecasting. By simplifying the model to focus on linear relationships within the data, Dlinear astonishingly surpasses more complex models in performance. This revelation underscores the potential of minimalistic approaches in extracting temporal relationships effectively, thereby questioning the necessity of complex architectures for certain forecasting tasks.

Moreover, the Modern Temporal Convolutional Network (ModernTCN) by Luo et al. (2024) \cite{luo2024moderntcn}, leverages convolutional operations to model temporal dependencies. ModernTCN employs dilated convolutions, residual connections, and layer normalization, resulting in a highly effective time series forecasting model with reduced computational overhead. However, ModernTCN may suffer from limited representation capabilities due to its lightweight backbone, potentially leading to inferior performance in certain tasks.

Additionally, SegRNN by Lin et al. (2023) \cite{lin2023segrnn} revisits RNN-based methods by introducing segment-wise iterations and parallel multi-step forecasting. These techniques reduce the number of recurrent steps and enhance forecasting accuracy and inference speed, making SegRNN a highly efficient tool for long-term time series forecasting. Despite these improvements, SegRNN can be constrained by the inherent limitations of RNNs, such as vanishing gradients and difficulty in capturing long-term dependencies.

Lastly, TiDE by Das et al. (2023) \cite{das2023long} combines the strengths of linear models and multi-layer perceptrons (MLPs). TiDE’s dense encoder architecture captures both linear and nonlinear dependencies in time series data, offering robust performance without the excessive computational demands of more complex models like transformers. Nonetheless, TiDE can struggle with modeling non-linear dependencies and may face challenges when dealing with complex covariate structures.

Our PIAD-SRNN model builds upon decomposition techniques used in Dlinear, Autoformer, and Fedformer, while significantly enhancing computational efficiency. By optimizing its architecture, PIAD-SRNN achieves faster training and inference times than leading transformer-based models, making it particularly suitable for real-time applications.

\section{Data Analysis}
\label{sec:app-data-analysis}

\begin{figure}[t!]
    \centering
    \subfigure[Indoor CO$_{2}$ Concentration (CO$_{2\_\text{in}}$)]{
        \includegraphics[width=0.48\textwidth]{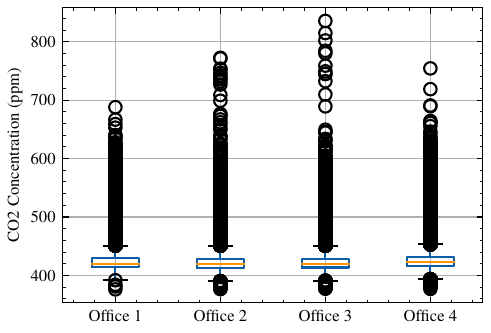}
        \label{fig:indoor_CO2}
    }
    \hfill
    \subfigure[Indoor Temperature ($T_{\text{in}}$)]{
        \includegraphics[width=0.48\textwidth]{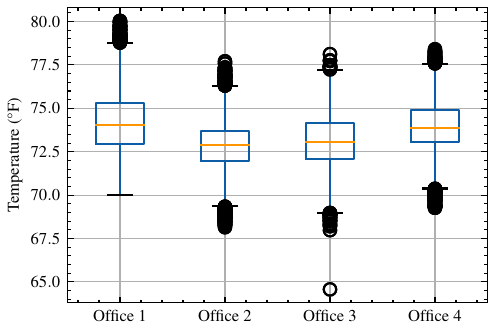}
        \label{fig:indoor_temp}
    }
    \hfill
    \subfigure[Outdoor Temperature ($T_{\text{out}}$)]{
        \includegraphics[width=0.48\textwidth]{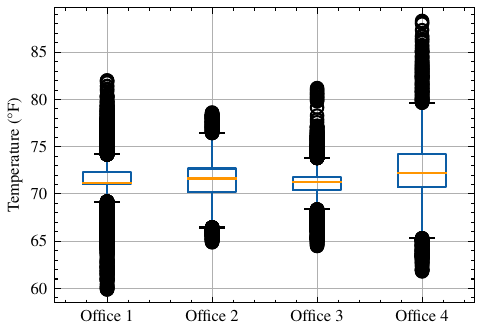}
        \label{fig:outdoor_temp}
    }
    \caption{Comparative environmental conditions across four
office spaces.}
 \vspace{-10mm}
    \label{fig:environmental_conditions}
\end{figure}

In Figure \ref{fig:environmental_conditions}, we present a comprehensive depiction of the environmental conditions across four different office settings through the analysis of indoor temperature, outdoor temperature, and indoor CO$_2$ concentration. 
The subfigure \ref{fig:indoor_CO2} focuses on the indoor CO$_2$ concentration levels. Across all offices, the median CO$_2$ concentration is consistent, indicating a general adherence to acceptable IAQ standards. However, the presence of numerous high outliers in each office suggests episodic events where the CO$_2$ levels exceed typical ranges. These instances could potentially be linked to variable occupancy levels, inconsistent operation of ventilation systems, or activities within the offices that temporarily increase CO$_2$ production.

The indoor temperature, illustrated in the subfigure \ref{fig:indoor_temp}, reveals a notable variance between offices. Office 1 exhibits a broad temperature range with several high outliers, indicative of sporadic deviations from the norm, which could be attributed to intermittent HVAC system performance or external factors affecting the indoor climate. In contrast, Offices 2 and 3 display a narrower interquartile range, suggesting a more consistent application of temperature control measures. Office 4, while also showing a relatively stable temperature range, has more outliers on the lower end, pointing to occasional dips in the indoor temperature.

The subfigure \ref{fig:outdoor_temp} captures the outdoor temperature conditions for each office location. Offices 1, 3, and 4 show a wider range of temperature values, which may reflect either a larger variance in outdoor climatic conditions or a more extended period of data capture. Office 2 demonstrates tighter temperature controls with minimal outliers, suggesting either less climatic variability or more effective mitigation of outdoor temperature fluctuations. Overall, the multitude of data eliminates the statistical biases, and the variety of the data supports the training of a robust model.

Analysis of the dataset reveals the presence of missing values, as detailed in Table \ref{tab:inputvar}, corresponding to the period from January 2020 to April 2020. The missing data is imputed, and appropriate sections of the dataset are selected for training, testing, and validation to maintain the integrity of our analysis. The upcoming section elaborates on this method and its significance in our research.

\end{document}